\documentclass[journal]{IEEEtran}

\usepackage{cite}

\usepackage{array}

\ifCLASSOPTIONcompsoc
 \usepackage[caption=false,font=normalsize,labelfont=sf,textfont=sf]{subfig}
\else
 \usepackage[caption=false,font=footnotesize]{subfig}
\fi

\usepackage{url}

\usepackage[section]{placeins}
\usepackage{graphicx}
\usepackage[export]{adjustbox}
\usepackage{tabulary}
\usepackage{tabularx}

\usepackage{makecell}

\newcolumntype{Z}{ >{\centering\arraybackslash}X }

\makeatletter 
\newcommand\HUGE{\@setfontsize\Huge{30}{40}}
\makeatother

\usepackage{amsfonts}
\usepackage{mathtools}
\usepackage{multirow}
\usepackage[skip=0.5\baselineskip]{caption}
\usepackage{float}
\floatstyle{plaintop}
\restylefloat{table}
\newcommand{\vect}[1]{\boldsymbol{#1}}

\begin{document}

\title{Word2Pix: Word to Pixel Cross Attention Transformer in
Visual Grounding}

\author{Heng Zhao, Joey Tianyi Zhou and Yew-Soon Ong,~\IEEEmembership{Fellow,~IEEE}
\thanks{Heng Zhao and Joey Tianyi Zhou are with Institute of High Performance Computing, A*STAR, Singapore 138632. (E-mail: azurerain7@gmail.com; joey.tianyi.zhou@gmail.com).}
\thanks{Yew-Soon Ong is with School of Computer Science and Engineering, Nanyang Technological University, Singapore, 639798. (E-mail: asysong@ntu.edu.sg).}}

\markboth{ }%
{Shell \MakeLowercase{\textit{et al.}}: Bare Demo of IEEEtran.cls for IEEE Journals}

\maketitle

\begin{abstract}
Current one-stage methods for visual grounding encode the language query as one holistic sentence embedding before fusion with visual feature. Such a formulation does not treat each word of a query sentence on par when modeling language to visual attention, therefore prone to neglect words which are less important for sentence embedding but critical for visual grounding. In this paper we propose Word2Pix: a one-stage visual grounding network based on encoder-decoder transformer architecture that enables learning for textual to visual feature correspondence via word to pixel attention. The embedding of each word from the query sentence is treated alike by attending to visual pixels individually instead of single holistic sentence embedding. In this way, each word is given equivalent opportunity to adjust the language to vision attention towards the referent target through multiple stacks of transformer decoder layers. We conduct the experiments on RefCOCO, RefCOCO+ and RefCOCOg datasets and the proposed Word2Pix outperforms existing one-stage methods by a notable margin. The results obtained also show that Word2Pix surpasses two-stage visual grounding models, while at the same time keeping the merits of one-stage paradigm namely end-to-end training and real-time inference speed intact.

\end{abstract}


\IEEEpeerreviewmaketitle

\section{Introduction}

\IEEEPARstart{I}n this paper we address the task of visual grounding, or referring expression comprehension, where the target is to localize the object instance in an image with a bounding box given a natural language expression which describes the referred object. From one perspective, image classification, object detection and image segmentation are fundamental vision recognition tasks. These supervised optimization tasks guide AI systems to establish correspondences between language tokens (class labels) and image feature from entire image, image regions and to image pixels respectively. However in visual grounding, models are expected to learn correspondence between object instances and sentence or phrases. These language queries may comprise many words that are not just class labels but can also be attribute words, location words or even more complicated relational descriptions.\\

\begin{figure}[!t]
\centering
\includegraphics[width=3.45in]{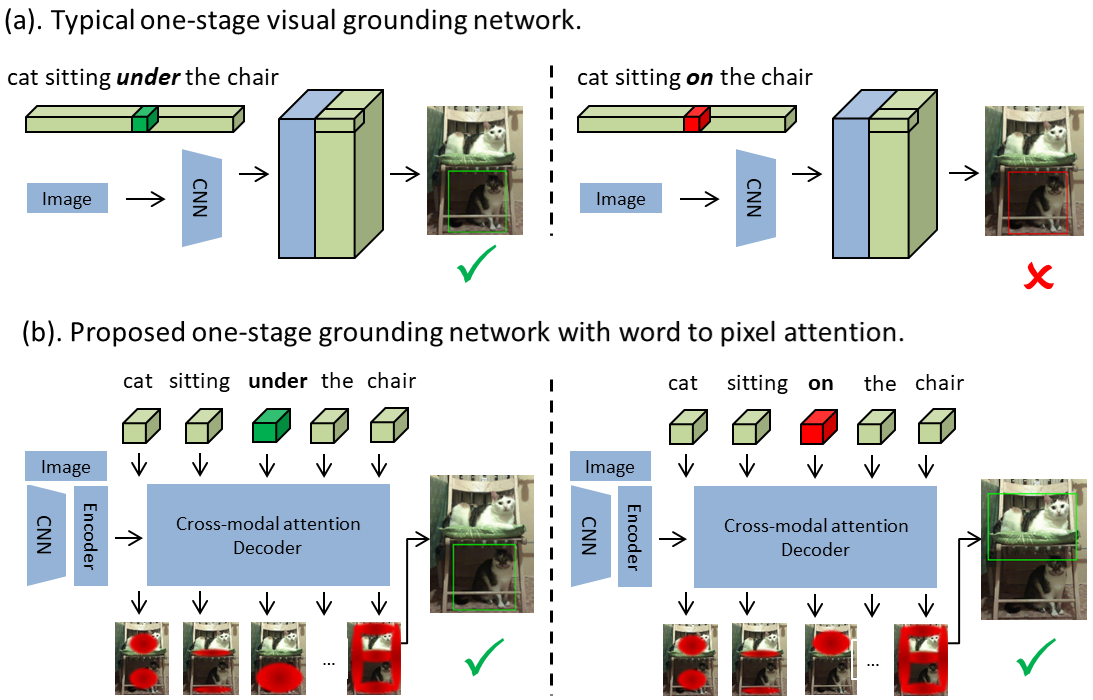}
\DeclareGraphicsExtensions.
\caption{Given a language query comprised of many words or tokens, visual grounding network is to predict a bounding box for the referred object. (a). Existing one-stage visual grounding networks encode the whole query sentence into one vector which is then duplicated and fused with visual feature at each spatial location. This formulation models language to visual attention at sentence level, therefore the impact of each individual word is not emphasized. (b). Word2Pix: our proposed one-stage grounding network that models word to pixel attention, offering a chance for each query word to attend to visual feature individually instead of being encoded into a holistic sentence embedding.}
\label{fig:intro}
\end{figure}

Existing methods for visual grounding can be split into two categories: two-stage and one-stage strategies. Two-stage visual grounding networks achieve state-of-the-art performance\cite{MATTNET}\cite{CM_Erase}\cite{refnms} with the capability of modeling various relationships between context objects in an explicit way. However, two stage methods suffer from slow inference time (usually at hundreds of milliseconds) and the performance is bound by the quality of region proposals from pre-trained detectors whose model weights cannot be optimized for visual grounding target during training. 

To improve inference efficiency and enable end-to-end training, many one-stage methods\cite{ZSG}\cite{FAOA}\cite{resc}\cite{RCCF} are gaining interest recently. The common strategy for these methods is to fuse visual and textual feature at middle stage and then formulate the localization as a regression problem. As shown in Fig. \ref{fig:intro}(a), language query is encoded into one sentence level vector embedding and then duplicated at each spatial location to facilitate the regression process after the multi-modal feature fusion. In this formulation, the language modality can only provide sentence level guidance for visual localization because the embedding for each word is implicitly merged as one single sentence vector. This is a disadvantage as individual word cannot directly interact with the visual feature so the importance of each word is diluted. For example in Fig. \ref{fig:intro}(a) where the query sentence is \textit{cat sitting under the chair}, the major component of the sentence embedding is taken by subject word (\textit{cat}), verb (\textit{sitting}) and object word (\textit{chair}). However, word indicating location and relationship (\textit{under}) is more critical in this case but it has to share the same sentence embedding with other words. As a result, the network fails to change its attention when the word \textit{under} is replaced with \textit{on} because individual word loses the chance to independently guide the network to focus on the corresponding visual feature. Additional qualitative examples are illustrated in latter section (Fig. \ref{fig:ones_comp1}).

Inspired by the success of transformers\cite{transformer} in NLP research and computer vision \cite{detr}, we propose Word2Pix: a one-stage visual grounding network based on transformer architecture to overcome the shortcomings of existing methods while at the same time maintaining the merits of one-stage paradigm. As illustrated in Fig. \ref{fig:intro}(b), the transformer decoder takes individual word token embedding as input, enabling each word to guide language to visual attention independently. This word to pixel attention provides a chance for each word in a sentence to impact the grounding results individually so that the network can learn to focus on critical word of the sentence without being biased towards the dominant words. Specifically, each word is able to attend to different visual pixels in first decoder layer and this attention is adjusted towards referent object via subsequent layers by the fusion with visual feature. Comparing with the two-stage methods, Word2Pix eliminates the need for proposal networks and enables end-to-end training, allowing gradient to reach visual feature extractor efficiently by connecting encoder layers directly to CNN backbone. Our proposed model is able to learn word feature with an embedding weight matrix or utilize off-the-shelf language models such as BERT\cite{bert}. At prediction stage, a final visual representation refined by word to pixel attention is used for object category classification, attribute learning and bounding box regression. Experiments on three referring expression datasets (RefCOCO\cite{referitgame}, RefCOCO+\cite{referitgame} and RefCOCOg\cite{refcocog}) show that our method greatly extends the performance of one-stage methods to surpass state-of-the-art two-stage ones, at the same time keeps the benefits of one-stage methods for real-time inference and end-to-end training.

We summarize our contributions in the following points:
\begin{itemize}

\item We propose Word2Pix: a one-stage visual grounding framework with word to pixel attention, enabling the network to consistently adjust the attention of each word on referent object and allowing a chance for individual word to guide the language-visual attention independently.

\item We are the first to formulate word to pixel attention via encoder-decoder transformer architecture for visual grounding, which improves the robustness and explainability of existing one-stage methods. The proposed method enables gradient to propagate efficiently through low level visual feature encoder and CNN backbone.

\item Our method extends the performance of existing one-stage by a margin to surpass two-stage methods, setting new state-of-the-art on three datasets and at the same time keeps real-time inference speed.
\end{itemize}

\section{Related Work}
\subsection{Visual Grounding}
Visual grounding algorithms can be categorized as two-stage and one-stage methods. The former localizes referent objects in two steps. \textit{1)} Extract visual features for candidate objects from off-the-shelf detectors. \textit{2)} Rank language query-object candidate pairs and select the candidate with the highest ranking score. Although in the second step, different modeling techniques can be exploited such as context modeling\cite{obj_retrvl}\cite{refcocog}\cite{refcoco}\cite{relationship_mod}\cite{context_var}\cite{cmrin}, modular attention\cite{MATTNET}\cite{CM_Erase}\cite{refnms}, graph and tree modeling\cite{lcmcg}\cite{dga}\cite{nmtree}, two-stage methods still have to rely on off-the-shelf object detectors\cite{frcnn}\cite{mrcnn} to extract visual features for candidate objects. As a result, two-stage methods suffer from performance bottleneck from proposal quality and slow inference speed. 

To address these disadvantages, many one-stage methods are proposed recently. They usually fuse textual and visual feature at middle stage and perform bounding box regression at each spatial location in training. Thus the training can be done in an end-to-end way and the inference speed at real-time. In general the fusion process is dense. For example, the language feature is duplicated and concatenated with visual feature at each spatial location\cite{FAOA}\cite{resc}\cite{ZSG}\cite{MCN}. Although the proposal generation and refinement step (NMS) in two-stage methods is not needed for one-stage methods, anchor box design at training stage is still needed for better performance. Alternatively, \cite{SSG}\cite{corrfilter} propose to predict the center and the size of the referred object without using anchor boxes in training. In order to efficiently fuse textual and visual feature, almost all one-stage methods encode the query sentence into single vector representation. As a result, one-stage methods are prone to neglect important information in a query sentence. To improve this situation, \cite{resc} propose to construct several sub-queries from input sentence to iteratively adjust the sentence embedding. However, since the embedding for each sub-query remains single vector, it is still subjected to possible neglect to important words. For example, a sub-query that focuses on a wrong target will propagate down to subsequent sub-queries, leading to the inability to attend to the critical word.

\subsection{Vision Transformer} 
Transformer architecture is originally proposed in \cite{transformer} for neural machine translation task in natural language processing. Inspired by the success of the self-attention mechanism, researchers have explored to use transformer framework for fundamental vision tasks such as image classification\cite{igpt}\cite{vit}, object detection\cite{detr}\cite{ddetr}\cite{tsp},segmentation\cite{setr} and other low-level vision tasks such as image generation, super-resolution and de-noising\cite{imgtrans}\cite{ipt}. \cite{vit} splits an image into patches and feeds them into a stack of transformer encoder to learn a representation for classification. \cite{igpt} leverages generative pre-training to learn visual representation in a self-supervised way with stacks of transformer decoder block. The encouraging work DETR proposed in \cite{detr} formulates object detection as a set prediction problem, eliminating the need for proposal generation, anchor box design and Non-Maximal Suppression (NMS) all at once. This work sets a new paradigm for object detection and inspired many following works such as deformable DETR\cite{ddetr} and Transformer-based Set Prediction (TSP)\cite{tsp}. \cite{ddetr} proposed deformable attention operation to solve the convergence issue in \cite{detr} while in \cite{tsp} an encoder-only transformer based detector is proposed.

\subsection{Multi-modal Transformer}
The encoder-decoder architecture of transformer can be conveniently adapted to multi-modal tasks such as captioning, question-answering, reasoning and visual grounding. VideoBERT proposed in \cite{videobert} learn joint video-text representation with transformers in a self-supervised way for downstream tasks. A multi-modal transformer is proposed in \cite{transforcap} for image captioning where object proposals generated from detectors are fed into encoder, while the decoder learns a language model conditioned by the encoder outputs. Language and vision pre-training is becoming a trend in this field\cite{visualbert}\cite{vlbert}\cite{vilbert}. 

In general, existing multi-modal transformers focus on generic learning for vision-language feature representations with various pre-training techniques on large scale dataset such as Visual Genome\cite{vgenome}, and then the model can be fine-tuned on downstream tasks such as referring expression comprehension. However, existing transformer based models take region proposal from off-the-shelf object detectors as input for visual feature and follow the two-stage paradigm on visual grounding task, where they suffer from the same shortcoming of slow inference speed and bottleneck from object detectors. In contrast, our method focus on visual grounding task without relying on pre-training on large scale dataset.

\section{Method}
In this work, we describe the proposed architecture in detail. Inspired by the success of self-attention and corss-attention mechanism used in transformers\cite{transformer} for NLP tasks, we model our word to pixel attention with transformer framework. To be specific, an encoder-decoder transformer is proposed where the encoder is directly connected to the CNN backbone for visual feature extraction and the decoder enables guidance from language to visual feature at word and pixel level respectively. In the following subsections we will review attention mechanism and its usage in transformers as preliminaries followed by our proposed architecture.

\begin{figure*}[!htb]
\centering
\includegraphics[height=3.5in]{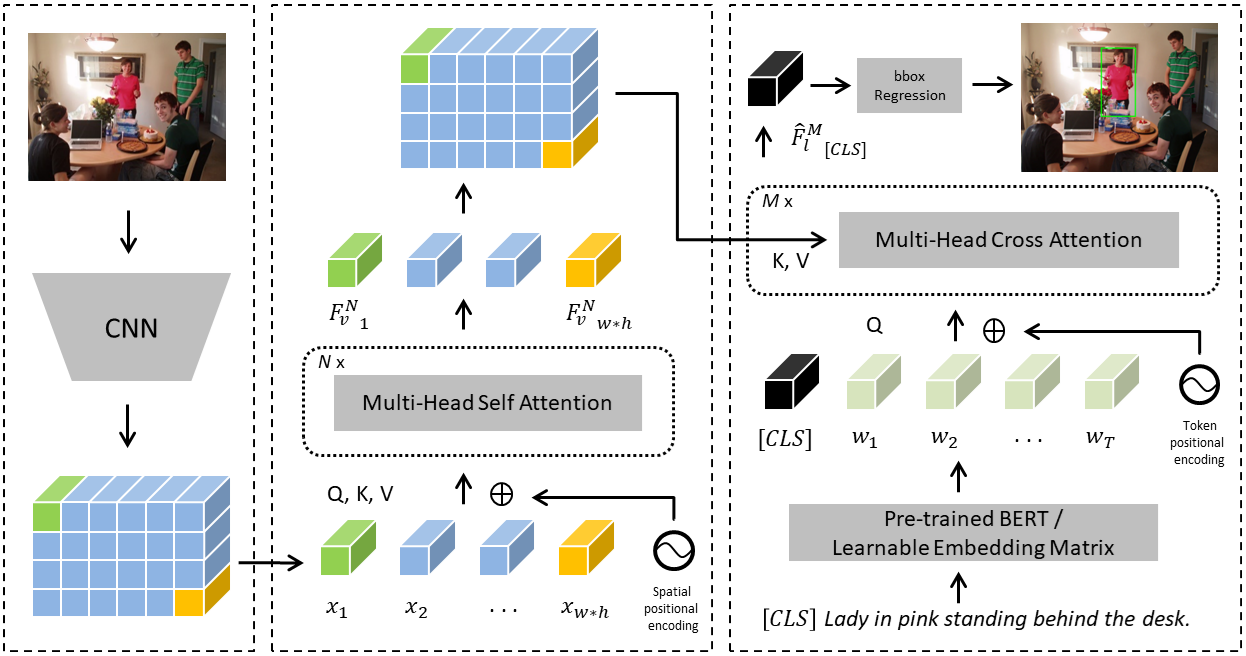}
\DeclareGraphicsExtensions.
\caption{Overview of Word2Pix: the proposed one-stage visual grounding network with transformer. Pixel level visual feature from backbone CNN are flattened as encoder input for refinement through self attention mechanism. Decoder takes a sequence of word token embeddings and \textit{[CLS]} embedding as input for queries where each query attends to the visual feature map at pixel level via multi-head cross attention. After $M$ stacks of cross attention layer the refined joint feature at \textit{[CLS]} token is taken for bounding box prediction.}
\label{fig:arch}

\end{figure*}

\subsection{Preliminaries: Multi-head Attention in Transformers}
\label{Preliminaries}
Attention mechanism is generally used to generate a new feature representation from original feature vectors which is specifically attended by another query feature. For instance, with given vector $Q$, $K$ and $V$ as query, key and value vector respectively, the attended vector is a weighted sum of value vector where the attention weights is calculated with dot product of query and key vector\cite{transformer}:
\begin{equation}
Attention(\vect{Q},\vect{K},\vect{V}) = softmax(\frac{\vect{Q}\vect{K}^\top}{\sqrt{d}})\vect{V}
\end{equation}
\label{eq:attn}
where $d$ is the dimension of vectors. The above attended vector is considered as a single-head attention and multiple single-head attention vectors are concatenated and scaled to form a multi-head attention. Note that the length of the attended output vector is aligned with the length of the query vector. Query can be from any source but key and value vectors are usually from the same origin. Specifically for self-attention in language representation learning, query, key and value are all from the same sentence. This enables the model to learn word vector embedding in a context-aware manner. Additionally in \cite{transformer}, a stackable transformer layer is formed by adding Layer Normalization (LN)\cite{ln} and an Multi-Layer Perceptron (MLP) after Multi-Head Self-Attention (MHSA) operation together with residual connections.
\\

\subsection{Word2Pix: One-stage Visual Grounding with Transformer}
We build our model with three parts: $(1)$ visual encoding branch $(2)$ word to pixel attention branch $(3)$ targets prediction head.
\\

\noindent \textbf{Visual Encoding Branch} The vision branch includes a CNN backbone from ResNet\cite{resnet} for general visual feature extraction and a stack of transformer encoder layers. 

Concretely, given an RGB image input with shape $H\times{W}\times{3}$, the image feature from the output of mid-layer CNN is of shape $\frac{H}{d}\times{\frac{W}{d}}\times{C}$ where $C$ is the number of feature maps. The feature maps are then flattened along spatial axis and transformed to a sequence of vectors $\{\vect{x_k}\}_{k=1}^{h\times{w}}$ where $h=\frac{H}{d}$, $w=\frac{W}{d}$ and $\vect{x_k}\in \mathbb{R}^D$. As a default, the dimension of feature maps from ResNet backbone is set to $C=2048$ and is reduced to $D=256$ via a linear layer. $d$ is set to $32$ to keep the length of the sequence at a reasonable number for the following self-attention based module. Due to the fact that the self-attention mechanism is pair-wise and ignores the order of the input sequence, positional encoding has to be added for the flattened visual feature. Thus the input visual feature for transformer encoder layer is denoted as:
\begin{equation}
\vect{F_v^0} = \{\vect{x_k}+ \textrm{PosEmb}_v(k)\}_{k=1}^{h\times{w}}
\end{equation}
\label{eq:enc_inp}
where $\textrm{PosEmb}_v(\cdot)$ is the sine positional encoding in \cite{detr}.

Feature after each transformer encoder layer can be denoted as:
\begin{equation}
\vect{F_v^{i+1}} = \vect{F_v^i}+\textrm{FFN}(\vect{F_v^i} + \textrm{MHSA}(\vect{F_v^i}))
\end{equation}
\label{eq:enc}
where $\textrm{FFN}(\cdot)$ is a 2-layer MLP and \textrm{MHSA} is the Multi-Head Self-Attention described in Section. \ref{Preliminaries}. Layer Normalization is omitted here for brevity. 
\\

\noindent \textbf{Word to Pixel Attention Branch} In this branch we model word to pixel attention: the correspondence between individual word tokens and pixel-level visual feature with transformer decoder. Specifically, given a natural language query sentence comprising of many words $\vect{S}=\{\vect{w_k}\}_{k=1}^T$, where $T$ is the length of the sentence and $\vect{w_k}\in\mathbb{R}^D$ is the embedding for each word. The embeddings of words can either be learned with an embedding matrix or obtained with a pre-trained language model. These word embeddings serve as the sequential query input $\vect{F_l^0}$ for decoder layer:

\begin{equation}
\vect{F_l^0}  =  \{\vect{w_k}+ \textrm{PosEmb}_l(k)\}_{k=1}^{T+1} 
\end{equation}
\label{eq:dec_inp}
where $\textrm{PosEmb}_l(\cdot)$ is a learnable positional embedding matrix with random initialization. In this way, the independence of each word is guaranteed and the sequential order of words are encoded via positional encoding. Note that this is in contrast to existing one-stage methods which consider holistic sentence embedding for feature fusion. For example in \cite{resc}, query sentence is encoded as $\vect{q}=\sum^{T}_{k=1}\alpha_k\vect{w_k}$ where $\vect{q}\in\mathbb{R}^D$ is the of the same dimension with each word embedding $\vect{w_k}$. This single vector is weighted sum of all word embeddings, thus the importance of individual words becomes diluted and the sequential order of words is lost.

After $M$ stacks of encoder layer, the output $\vect{F_v^M}=\{\vect{\hat{x}_k}\}_{k=1}^{h\times{w}}$ can be considered as a spatial-context-aware pixel vector sequence. With this vector sequence as the visual memory for an input image, we build the word to pixel attention between self-attended word embedding sequence $\vect{\hat{F}_l^i}=\textrm{MHSA}(\vect{F_l^i})=\{\vect{\hat{w}_k}\}_{k=1}^{T+1}$ and pixel feature sequence $\vect{F_v^M}=\{\vect{\hat{x}_k}\}_{k=1}^{h\times{w}}$ in each decoder layer. Self-attention is applied to the word embedding sequence to generate context aware feature, followed by word to pixel cross-attention: 

\begin{equation}
\begin{array}{rcl}
\vect{\hat{F}_l^i} &=& \vect{F_l^i} + \textrm{MHSA}(\vect{F_l^i}) \\\\ 
\vect{F_l^{i+1}} &=& \vect{F_l^i}+\textrm{FFN}(\vect{F_l^i} + \textrm{MHCA}(\vect{\hat{F}_l^i}, \vect{F_v^M}))
\end{array}
\end{equation}
\label{eq:dec}

The Multi-Head Cross Attention (MHCA) for each head is computed as:
\begin{equation}
\begin{array}{rcl}
\textrm{MHCA}_{head}(\vect{\hat{F}_l^i},\vect{F_v^M}) &=& softmax(\frac{\vect{Q_l}\vect{K_v}^\top}{\sqrt{d_{dim}}})\vect{V_v}\\\\
\vect{Q_l} &=& \vect{W_Q}^\top \vect{\hat{F}_l^i}\\\\
\vect{K_v} &=& \vect{W_K}^\top \vect{F_v^M}\\\\ 
\vect{V_v} &=& \vect{W_V}^\top \vect{F_v^M}\\ 
\end{array}
\end{equation}
\label{eq:mhca}
where $\vect{W_Q}$, $\vect{W_K}$ and $\vect{W_V}$ are learnable weight matrices for query, key and value respectively. Note the cross-attention weight is of shape $(w\times{h})\times(T+1)$ where each textual token is attending to all pixel locations independently. 
\\

\noindent \textbf{Targets Prediction} Instead of only using bounding box regression as the optimization target, our grounding model also predicts the category and attributes of the referred object with the same high-level feature with rich semantic information. The source of value and key vectors in Multi-Head Cross-Attention block for every layer of the decoder is the memory visual feature from the encoder, and the output is a weighted sum of this memory feature which is also from visual space. Thus the semantics of this visual feature at the last decoder layer can naturally be used for all three vision-related prediction tasks. We use the visual representation attended by the dummy token \textit{[CLS]} for all prediction heads.

A two layer MLP with ReLU activation is constructed for bounding box regression predicting the center coordinates, height and width of the target object. The classification heads for category and attributes prediction are two simple linear layers followed by sigmoid function respectively.
\\

\noindent \textbf{Optimization Objectives} We optimize the proposed grounding model with four losses:
\begin{equation}
\begin{array}{cr}
\mathcal{L} = \mathcal{L}_{ce} + \lambda_{bce}\mathcal{L}_{bce} + \lambda_{l_1}\mathcal{L}_{l_1} + \lambda_{giou}\mathcal{L}_{giou}
\end{array}
\end{equation}
\label{eq:loss0}
specifically, given three ground truth labels: class label $\vect{Y}=\{{Y_i}\}_{i=1}^{N_c}$, attribute labels $\vect{y}=\{{y_i}\}_{i=1}^{N_a}$ and location bounding box coordinates $\vect{b}$ of the referent object, the four losses are noted as follows:
\begin{align*}
\mathcal{L}_{ce} =  -\frac{1}{N_c}\sum^{N_c}_{i=1}Y_ilog(\hat{Y}_i) 
\end{align*}
\label{eq:loss1}
\begin{equation}
\mathcal{L}_{l_1} = \left\|\vect{b}-\hat{\vect{b}}\right\|_1
\end{equation}
\label{eq:loss2}
\begin{align*}
\mathcal{L}_{bce}=-&\frac{1}{|\{i:y_i=1\}|}\sum^{}_{\{i:y_i=1\}}log(\hat{y}_i)\\
-&\frac{1}{|\{i:y_i=0\}|}\sum^{}_{\{i:y_i=0\}}log(1-\hat{y}_i)
\end{align*}
\label{eq:loss3}
where $N_c$ and $N_a$ is the number of categories and attributes respectively, $\hat{Y_i}$ is the softmax of predicted logits for class label $i$. $\vect{\hat{b}}$ denotes the predicted bounding box coordinates. The generalized IoU loss $\mathcal{L}_{giou}$ follows \cite{giou}. For multi-label prediction, we use a dynamically weighted binary cross-entropy loss where $|\{i:y_i=1\}|$ denotes the number of positive attribute labels that one object has.

\section{Experiments}
\subsection{Datasets}
We conduct our experiments on three datasets: RefCOCO\cite{refcoco}, RefCOCO+\cite{refcoco} and RefCOCOg\cite{refcocog} where the images are mostly from MSCOCO\cite{coco}. There are 19,994 / 19,992 / 25,799 images with 50,000 / 49,856 / 49,822 referred entities and 142,210 / 141,564 / 95,010 expressions in RefCOCO, RefCOCO+ and RefCOCOg respectively. The average number of words for expressions in RefCOCO and RefCOCO+ is $3.5$ while sentence in RefCOCOg is longer with $8.4$ words on average. Comparing with RefCOCO, expressions in RefCOCO+ are not allowed to contain words indicating absolute location of the referred object.

In terms of evaluation, the test set is split into $testA$ and $testB$ for RefCOCO and RefCOCO+ where $testA$ focuses on person while $testB$ focuses on object. There are two splits for RefCOCOg and we follow the split RefCOCOg-umd used in \cite{refcoco} where images in training, validation and testing sets do not overlap.

\subsection{Implementation and Training Details}
The images are resized with shorter side as $640$ with a maximum longer side as $1333$. The original aspect ratio is maintained and zero valued pixels are padded for batch training. No other data augmentation techniques are used during training. We choose ResNet-101 as our feature extractor and high level layers are discarded. $6$ layers of encoders are stacked on top of the backbone CNN with Layer Normalization applied after the attention operation. For language branch, the maximal input length is set to $10$ and $15$ for decoder. Dummy token \textit{[CLS]} is added at the beginning of the input sentence and \textit{[PAD]} is appended at the end to support batch training. Words are tokenized as indices in vocabulary and embedded with a learnable matrix where the vocabulary size of all expressions in RefCOCO is around $2000$. In case of embedding from BERT is used, a linear layer is added to project the dimension from $768$ to $256$ and the embedding from the dummy token \textit{[CLS]} is used as the feature for the whole sentence. Note that the BERT tokenizer is a sub-word tokenizer, thus the tokenized sentence length will be longer than the length of textual expression. We trim the token embedding sequence if the length exceeds the maximal input length of the decoder. 

For training objectives, there are $80$ categories for classification loss and $50$ attributes for binary cross entropy loss. The ground truth attribute labels is very sparse where the value of most labels is zero for one object, thus the attribute loss is dominated by negative samples. We implement a mean averaged attribute loss to balance the positive and negative labels. 

\begin{table*}[!htbp]
\resizebox{\textwidth}{!}{
\centering
\renewcommand{\arraystretch}{1.3}
\begin{tabular}{c|c|c|c|ccc|ccc|cc|c}
\hline
&
 \multirow{2}{*}{Methods}&
 \multirow{2}{*}{Venue} & 
 \multirow{2}{*}{Backbone}&  
 \multicolumn{3}{c|}{RefCOCO}&
 \multicolumn{3}{c|}{RefCOCO+}&
 \multicolumn{2}{c|}{RefCOCOg}& 
 \multirow{2}{*}{Time(ms)}\\ 
  
& & &  & val & testA(\%) & testB(\%) & val & testA(\%) & testB(\%) & val(\%) & test(\%) &   \\ 
\hline
\hline
\parbox[t]{2mm}{\multirow{8}{*}{\rotatebox[origin=c]{90}{\textbf{Two-stage} }}}
& ParalAttn\cite{paralattn}   & \textit{CVPR'18} & vgg16 & -- & 75.31 & 65.52 & -- & 61.34 & 50.86 &--&--&--\\

& NMTree\cite{nmtree} & \textit{ICCV'19} & vgg16 & 71.65 & 74.81 & 67.34 & 58.00 & 61.09 & 53.45 & 61.01 & 61.46 & --\\

& LGRAN\cite{lgran} & \textit{CVPR'19} & vgg16 & -- & 76.60 & 66.40 &--& 64.00 & 53.40 &--&--&--\\

& DGA\cite{dga} & \textit{ICCV'19} &   
vgg16 & -- & 78.42 & 65.53 &--& 69.07 & 51.99 &--& 63.28 &341\\

& RvGTree\cite{rvgtree} & \textit{TPAMI'19} & ResNet-101 & 75.06 & 78.61 & 69.85 & 63.51 & 67.45 & 56.66 & 66.95 & 66.51 & --\\

& MAttNet\cite{MATTNET} & \textit{CVPR'18} & ResNet-101 & 76.65 & 81.14 & 69.99 & 65.33&  71.62 & 56.02 & 66.58 & 67.27 & 320\\

& VL-BERT\cite{vlbert} & \textit{ICLR'20} & ResNet-101 & -- & -- & -- & 66.03 & 71.87 & 56.13 & -- & -- & --\\

& CMAE\cite{CM_Erase} & \textit{CVPR'19} & ResNet-101 & 78.35 & 83.14 & 71.32 & 68.09 & 73.65 & 58.03 & 67.99 & 68.67 & --\\

& Ref-NMS\cite{refnms} & \textit{AAAI'21} &   ResNet-101 & 80.70 & 84.00 & 76.04  & 68.25& 73.68 & 59.42 & 70.55 & 70.62 \\

& ViLBERT\cite{vilbert} & \textit{NIPS'19} & ResNet-101 & -- & -- & -- & 68.61 & 75.97 & 58.44 & -- & -- & --\\

\hline
\hline
\parbox[t]{2mm}{\multirow{5}{*}{\rotatebox[origin=c]{90}{\textbf{One-stage} }}}
& SSG\cite{SSG} & \textit{arXiv'18} &  Darknet-53 & -- & 76.51 & 67.50 & -- & 62.14 & 49.27 & 58.80 & -- & 25\\

& FAOA\cite{yolo3vg} & \textit{ICCV'19} &  Darknet-53 & 72.05 & 74.81 & 67.59 & 56.86 & 61.89 & 49.46 & 59.44 & 58.90 & 38\\

& ReSC-Large\cite{resc} & \textit{ECCV'20} &  Darknet-53 & 77.63 & 80.45 & 72.30 & 63.59 & 68.36 & 56.81 & 67.30 & 67.20 & 36 \\

& RCCF\cite{RCCF} & \textit{CVPR'20} &   DLA-34\cite{DLA}    & -- & 81.06 & 71.85 & -- &  70.35 & 56.32 & -- & 65.73 & 25\\
\cline{2-13}
& Word2Pix(ours) & -- & ResNet-101 &\textbf{81.20} &  \textbf{84.39} & \textbf{78.12} & \textbf{69.46}&  \textbf{76.81} & \textbf{61.57} & \textbf{70.81} & \textbf{71.34} & 29\\
\hline
\end{tabular}
\caption{Comparing with state-of-the-art methods on RefCOCO\cite{refcoco}, RefCOCO+\cite{refcoco} and RefCOCOg\cite{refcocog} datasets.}
\label{tab:res1}}
\end{table*}

In the current study, for the sake of reproducibility we considered a transformer encoder that is built on top of the DETR\cite{detr} and initialize the corresponding weights from models pre-trained on MSCOCO\cite{coco}. Note that all validation and testing images in RefCOCO, RefCOCO+ and RefCOCOg are also included in MSCOCO training dataset. For fair comparison, we re-train a model from scratch excluding all validation and testing images from grounding dataset for $100$ epochs. The model is optimized with AdamW\cite{adamw} with a batch size of $6$, the initial learning rate is set to $10^{-4}$ and $10^{-5}$ for transformer and CNN backbone respectively. Learning rate is decayed by $10$ at an epoch of $160$ until training stops at epoch of $180$. Due to the fact that the encoder is initialized from pre-trained weights while the decoder is randomly initialized. We freeze the CNN backbone and encoder weights during the first $80$ epochs. The weight coefficient for $\mathcal{L}_{ce}$, $\mathcal{L}_{bce}$, $\mathcal{L}_{l_1}$ and $\mathcal{L}_{giou}$ is set to $1$, $10$, $5$ and $2$, respectively.

\subsection{Comparison with State-of-the-art Methods}
We report our grounding results with comparison to existing one-stage and two-stage methods, as shown in Table \ref{tab:res1}. Following the one-stage paradigm, we observed significant improvements ($>5\%$ on average) over the state-of-the-art one-stage methods on three datasets. As mentioned previously, all existing one-stage methods uses one or several sentence level embedding for language query, hence the attention interaction between word tokens and visual feature have been neglected. In contrast, our Word2Pix model emphasizes attention from individual word to pixel level visual feature, bringing a large performance leap of $\sim6.5\%$ on \textit{testA} set of RefCOCO+. Built on top of \cite{FAOA}, \cite{resc} has been reported to offer better performance ($67.20\%$) over the former on the RefCOCOg test set where the average query sentence length is longer. Nonetheless, Word2Pix model is shown to extend the performance of those reported in \cite{resc} by a good margin of $\sim4\%$ on the same set, thus illustrating the importance of word to pixel attention, hence the benefits of our proposed approach.

We also compare Word2Pix with current state-of-the-art two-stage method\cite{refnms}. We observe a performance boost of $\sim2\%$ and $\sim3\%$ on the \textit{testB} set of RefCOCO and \textit{testA} of RefCOCO+ datasets, respectively. Small gains on validation and test set of other datasets are also observed.
When comparing with transformer based two-stage multi-modal methods\cite{vlbert}\cite{vilbert}, we reference their results obtained without additional training data for fair comparison. Aligned with other one-stage methods, our model is able to perform real-time inference while two-stage methods must make a trade-off between accuracy and inference speed due to the need for the proposal refinement post-processing step (NMS). As a matter of fact, recent state-of-the-art two-stage grounding models\cite{CM_Erase}\cite{refnms} share the same framework with MAttNet\cite{MATTNET} where the visual extractor is pre-trained mask-RCNN\cite{mrcnn} and the grounding module is modular attention from MAttNet\cite{MATTNET}. In all three methods, the visual extractor is not trained for grounding tasks. In contrast, our proposed one-stage model is able to tune visual extractor with grounding supervision signals in an end-to-end way. We show that tuning visual feature extractor is important for us to beat all two-stage methods in ablation experiments.

We note that the maximum number of input tokens from query sentence and computational efficiency remains a trade-off to be considered. In the present study, the number of input word tokens is set at $15$ for the experiments since the majority of the sentences do not exceed this number. On RefCOCOg despite some of the long sentences are trimmed, Word2Pix is shown to remain outperforming state-of-the-art one-stage methods who use holistic sentence embedding by $\sim4\%$.

\subsection{Ablation Studies}
In this subsection, we show evidence that our proposed end-to-end network is able to overcome the shortcomings of traditional one-stage and two-stage methods by several designed ablation experiments on RefCOCO\cite{refcoco} dataset. \\

\noindent \textbf{Word to Pixel vs Sentence to Pixel} Current state-of-the-art one-stage methods encode the holistic sentence query into one\cite{SSG}\cite{FAOA}\cite{corrfilter} or several\cite{resc} single vector embedding for textual and visual feature fusion. In this ablation experiment we demonstrate that the proposed word to pixel attention is superior to sentence to pixel attention which uses multiple holistic sentence level representations. Specifically, three experiment settings are designed: \textit{1)} replace decoder query inputs where each query token is a single vector sentence embedding from the pre-trained BERT model\cite{bert}. We take vector embedding of \textit{[CLS]} token from last $4$ layers of pre-trained BERT model as sentence level embedding, resulting $4$ different sentence level embeddings for a given language input. At last layer of the decoder, output feature vectors of the $4$ sentence level embeddings are pooled and forwarded to the prediction head. \textit{2)} instead of using pre-trained BERT feature for word token embedding, a word embedding matrix is learned with the dataset's vocabulary during training. \textit{3)} word and \textit{[CLS]} token embeddings are obtained from the last layer of pre-trained BERT model. The results are shown in Table \ref{tab:ablation2}. We observe only using sentence level embeddings for grounding task yields inferior performance, even comparing to model with word embedding learnt directly from dataset's vocabulary. We note that the vocabulary size for all language queries in RefCOCO dataset is only $2,000$, therefore the language embedding learnt from dataset is weak comparing to that from a pre-trained BERT model. This demonstrate the significance of word-level modeling and the superiority of our proposed framework.

\begin{table}[!htbp]
\resizebox{\columnwidth}{!}{
\centering
\begin{tabular}{c|ccc}
\hline
 \multirow{2}{*}{Decoder query input}&
 \multicolumn{3}{c}{RefCOCO}\\ 
& val & testA(\%) & testB(\%) \\ 
\hline
Sentence embedding(BERT) only  & $78.06$ &$81.28$ & $73.70$ \\
Word embedding(Learnt)  & $78.92$ & $82.18$ &$76.55$  \\
Word embedding(BERT) & $\textbf{81.20}$ & $\textbf{84.39}$ &$\textbf{78.12}$ \\
\hline
\end{tabular}
\caption{Word to pixel vs sentence to pixel in vision-language cross-attention.}
\label{tab:ablation2}}
\end{table}

\noindent \textbf{Off-the-shelf Visual Feature vs End-to-end Training} One key disadvantage of traditional two-stage methods is that the visual feature extractors or detectors cannot be fine-tuned with visual grounding supervision signals. Instead, their weights are fixed during the training of grounding networks. However, visual features suitable for object classification and detection task may need to be refined for language guided visual grounding task. For example, when an object detector is to decide whether a region of interest is a \textit{person} or not, features for action, color and pose wouldn't contribute much to this category classification task. However, these features are critical for visual grounding and they are usually encoded in lower layer of CNN backbone. Unable to fine-tune the low level visual extractor leads to inferior performance when comparing to those models who are able to be trained in an end-to-end way. To validate this, we initialize the CNN backbone and encoder of our model with weights pre-trained on COCO\cite{coco} object detection dataset (excluding RefCOCO\cite{refcoco}, RefCOCO+\cite{refcoco} and RefCOCOg\cite{refcocog} validation and testing set images). Experiments on RefCOCO dataset with the following settings are conducted. \textit{1)} freeze CNN backbone and encoder weights and train word to pixel attention decoder only. \textit{2)} freeze CNN backbone only. \textit{3)} end-to-end training with all model weights trainable. The results are shown in Table \ref{tab:ablation1}. It can be observed uniform improvements are obtained on all test and validation sets of RefCOCO\cite{refcoco} when more visual feature extractor weights are trainable. We observe a consistent improvement of $\sim5\%$ on \textit{testA} set where the referent objects all belong to \textit{person} category. This demonstrates that training visual feature extractor in an end-to-end way is especially critical for referring to objects with class-agnostic features. We note that freezing CNN backbone and encoder weights means the feature from off-the-shelf detectors are used for word to pixel attention. In this case, the performance of Word2Pix model ($75.58\%$) did not drop as much on \textit{testB} set when compared to end-to-end training ($78.12\%$). This is due to the fact that in \textit{testB}, referents are common objects other than \textit{person} category and the number of same type referent objects in one image is fewer comparing to \textit{testA} set. In this case a pre-trained class-aware object detector will contribute a good part for the accuracy.

\begin{table}[!htbp]
\resizebox{\columnwidth}{!}{
\centering
\begin{tabular}{c|ccc}
\hline
 \multirow{2}{*}{Training scheme}&
 \multicolumn{3}{c}{RefCOCO}\\ 
& val & testA(\%) & testB(\%) \\ 
\hline
Freeze pre-trained CNN \& encoder  & $75.26$ &$75.48$ & $75.58$ \\
Freeze Pre-trained CNN  & $76.40$ & $80.20$ &$75.74$  \\
End-to-end training  & $\textbf{81.20}$ & $\textbf{84.39}$ &$\textbf{78.12}$ \\
\hline
\end{tabular}
\caption{Effectiveness of end-to-end training vs pre-trained low-level visual extractors.}
\label{tab:ablation1}}
\end{table}

\noindent \textbf{Ablations on network architecture and losses} We further conduct ablation experiments with different number of decoder layers and different supervision signal combinations, as shown in Table \ref{tab:ablation3}. For fast verification purpose, we freeze the CNN backbone and encoder weights and only train decoder from random initialization. Thus the baseline model here with all four losses and number of decoder layer ($N=3$) suffers minor performance drop ($<1\%$) compared to the baseline model trained in an end-to-end way (main results in Table \ref{tab:res1}). We note that the impact of attribute and classification loss is not significant comparing to the model trained only with bounding box regression. We argue it is because the fact that in RefCOCO dataset less than one third of all referent objects have attribute labels while the rest have nothing to be trained with.

\begin{table}[!htbp]
\begin{center}
\begin{tabular}{ccc|ccc}
\hline
 \multicolumn{3}{c|}{Losses}& 
 \multicolumn{3}{c}{RefCOCO}\\ 
$\mathcal{L}_{l_1+l_{giou}}$& $\mathcal{L}_{ce}$& $\mathcal{L}_{bce}$& val & testA(\%) & testB(\%) \\ 
\hline
\checkmark & & & $80.05$ &$83.33$ & $76.78$ \\
\checkmark & \checkmark &  & $79.63$ &$83.07$ & $76.98$ \\
\checkmark & & \checkmark & $79.95$ &$83.45$ & $77.11$ \\
\checkmark & \checkmark & \checkmark & $\textbf{80.20}$ &$\textbf{83.60}$ & $\textbf{77.68}$ \\
\hline
\multicolumn{3}{c}{Number of decoder layers}&&&\\
\hline
\multicolumn{3}{c}{$N=1$}  & $77.58$ &$81.16$ & $74.80$ \\
\multicolumn{3}{c}{$N=2$}  & $79.06$ &$82.52$ & $75.72$ \\
\multicolumn{3}{c}{$N=3$}  & $\textbf{80.20}$ &$\textbf{83.60}$ & $\textbf{77.68}$ \\
\multicolumn{3}{c}{$N=4$}  & $80.07$ &$83.15$ & $77.29$ \\

\hline
\end{tabular}
\end{center}
\caption{Ablation on loss settings and number of decoder layers.}
\label{tab:ablation3}
\end{table}

\subsection{Qualitative Analysis}

\begin{figure*}[!htb]
\centering
\resizebox{\textwidth}{!}{
\begin{tabulary}{\linewidth}{c c c c c c c}
 & 
\textit{\thead{Q:cat sitting\\ \textbf{under} the chair}} & 
\textit{\thead{Q:cat sitting\\ \textbf{on} the chair}}& 
\textit{\thead{Q:man in \textbf{black} \\suit holding a hat}}&
\textit{\thead{Q:man in \textbf{gray} \\suit holding a hat}}&
\textit{\thead{Q:person on skateboard \\behind the \textbf{dog}}}&
\textit{\thead{Q:person on skateboard \\behind the \textbf{woman}}}\\\\
\Huge FAOA\cite{FAOA} &
\raisebox{-.5\height}{\includegraphics{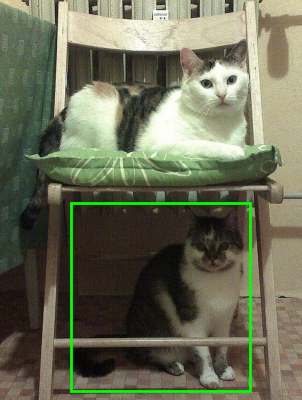}} & \raisebox{-.5\height}{\includegraphics{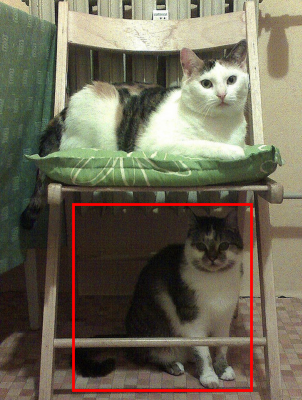}} &
\raisebox{-.5\height}{\includegraphics{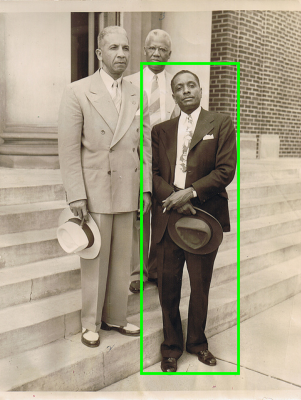}} & \raisebox{-.5\height}{\includegraphics{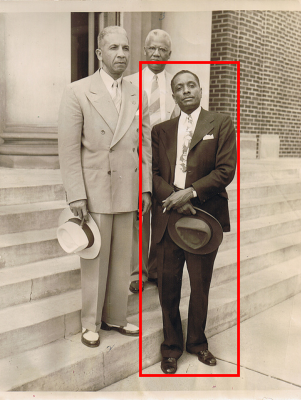}} &
\raisebox{-.5\height}{\includegraphics{comp/comp0/3_1.png}} & \raisebox{-.5\height}{\includegraphics{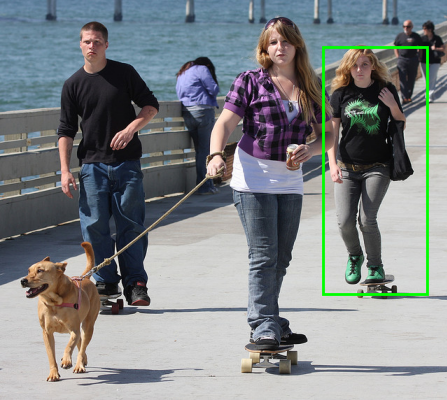}} \\\\
\Huge ReSC\cite{resc} & 
\raisebox{-.5\height}{\includegraphics{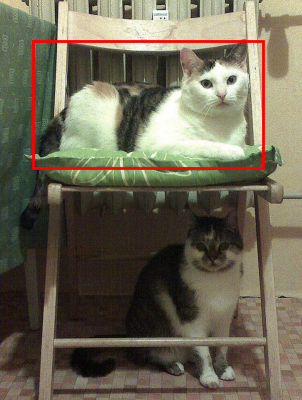}} & \raisebox{-.5\height}{\includegraphics{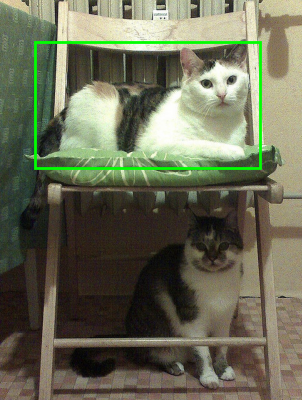}} &
\raisebox{-.5\height}{\includegraphics{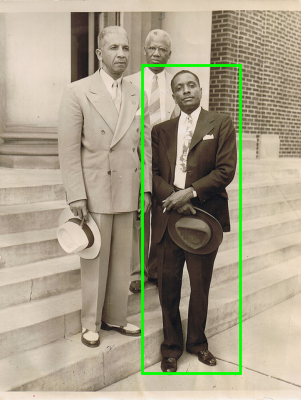}} & \raisebox{-.5\height}{\includegraphics{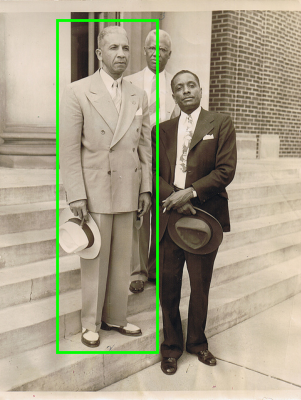}} &
\raisebox{-.5\height}{\includegraphics{comp/comp0/3_3.png}} & \raisebox{-.5\height}{\includegraphics{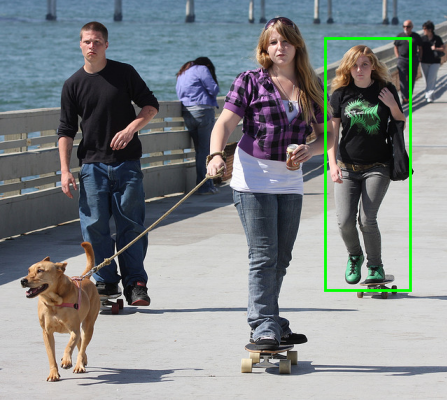}} \\\\
\Huge Word2Pix & 
\raisebox{-.5\height}{\includegraphics{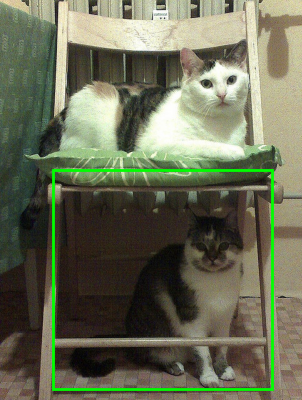}} & \raisebox{-.5\height}{\includegraphics{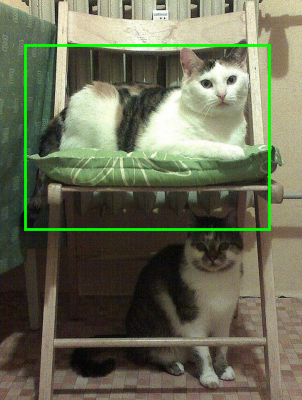}} &
\raisebox{-.5\height}{\includegraphics{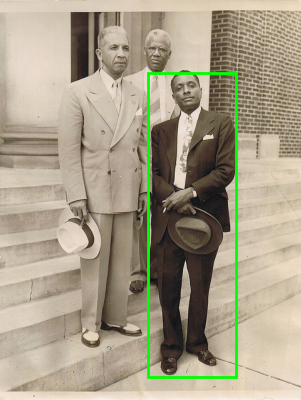}} & \raisebox{-.5\height}{\includegraphics{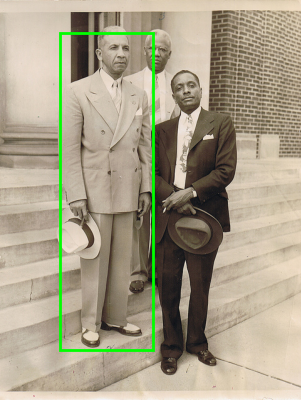}} &
\raisebox{-.5\height}{\includegraphics{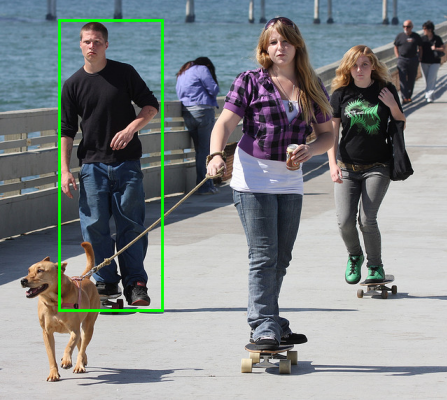}} & \raisebox{-.5\height}{\includegraphics{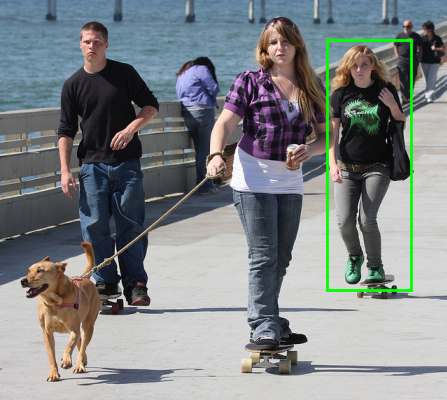}} \\\\
\end{tabulary}
}
\caption{Qualitative comparison with stage-of-the-art one-stage methods with similar (in vector embedding space) query sentences on same images with free text query input. The cosine similarity between the two similar sentence embeddings from pre-trained BERT model for the three input text pairs is $0.986$, $0.983$ and $0.977$ respectively. First row shows that FAOA\cite{FAOA} fails to detect subtle but critical word change in query input and generates the same grounding results. ReSC\cite{resc} uses several sentence level embeddings through multiple rounds of interaction with visual feature. In contrast, Word2Pix treats each word individually when attending to visual feature. Consequently, the model is able to learn to focus on critical word with a single forward pass and the robustness is improved. Bounding box color denotation: green for correct results and red for incorrect results.}
\label{fig:ones_comp1}
\end{figure*}

\begin{figure*}[!htb]
\centering
\resizebox{\textwidth}{!}{
\begin{tabulary}{\linewidth}{c c c }
\includegraphics[height=4.5cm]{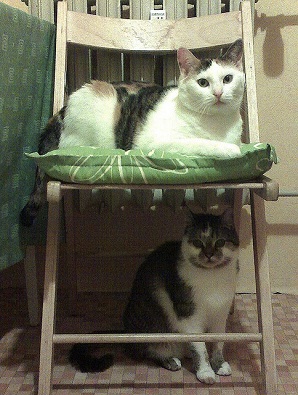} &
\includegraphics[height=4.5cm]{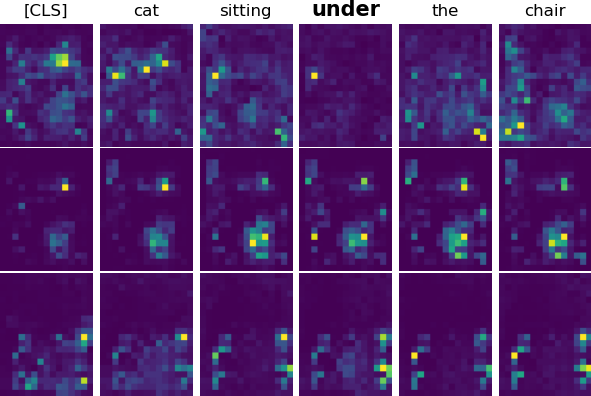} &
\includegraphics[height=4.5cm]{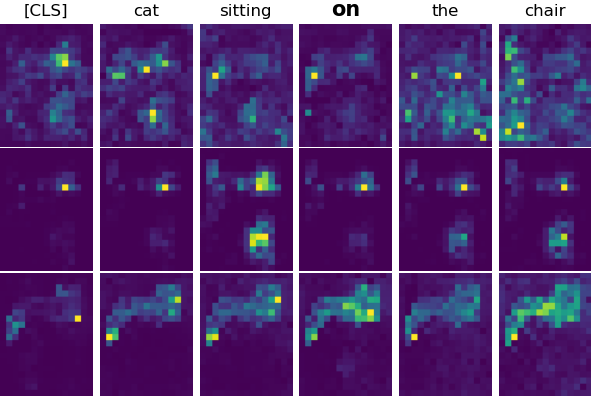}\\
\includegraphics[height=4.5cm]{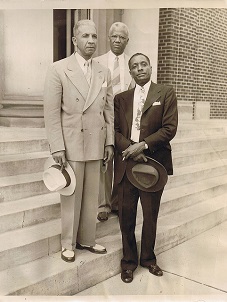} &
\includegraphics[height=4.5cm]{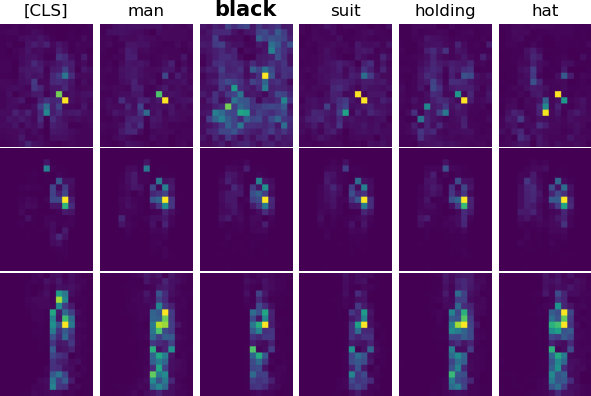} &
\includegraphics[height=4.5cm]{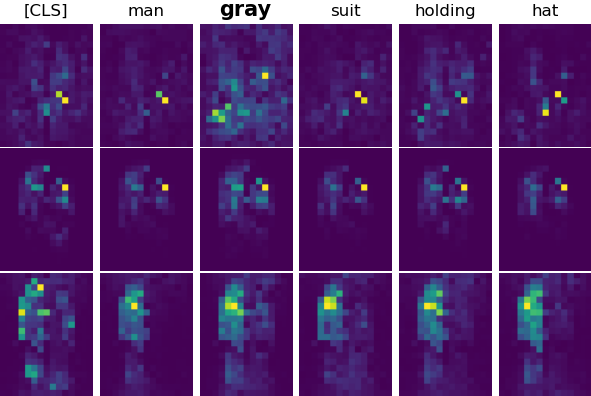}\\
\includegraphics[width=3.6cm]{amap/im_person.jpg} &
\includegraphics[width=6.8cm]{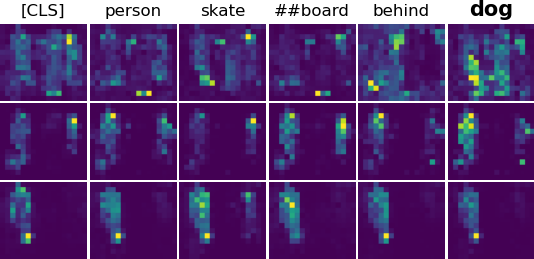} &
\includegraphics[width=6.8cm]{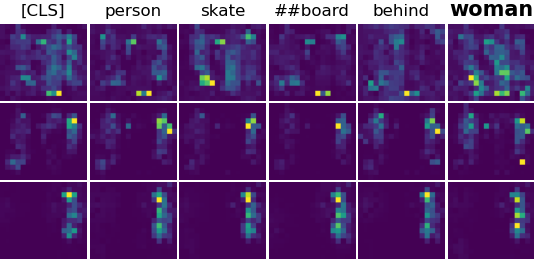}
\end{tabulary}
}
\caption{Cross attention weights between word tokens and pixel level visual feature. For each image, two similar queries are tested with a different keyword shown in bold. For the three sub-rows of each row, first and last sub-row correspond to first and final decoder layer attention weights respectively. Brighter pixels represent higher attention values.}
\label{fig:amap}
\end{figure*}

We illustrate that our proposed method is able to overcome the disadvantage of one-stage methods by qualitative examples in Fig. \ref{fig:ones_comp1}. We feed free-text input (query sentence that is free input from user instead of directly quoting from the dataset) to FAOA\cite{FAOA}, ReSC\cite{resc} and Word2Pix. Specifically, three pairs of similar queries are used to illustrate the weakness of using holistic sentence embedding when interacting with visual feature for visual grounding. We calculate the cosine similarities between the sentence embeddings of these three query pairs to show that the vector distance between holistic sentence embeddings is small (i.e., $0.986$, $0.983$ and $0.977$ respectively). It can be seen FAOA fails to distinguish the differences between input sentences and ReSC is able to improve but is still not robust against subtle but critical changes in query sentences. As a comparison, Word2Pix is able to detect minor changes in queries because each word in a query is able to affect the final grounding results individually. For example in the first two columns in Fig. \ref{fig:amap}, the embedding for word \textit{under} and \textit{on} contribute only a small part in the whole sentence embedding but they are critical for referring to the correct referent. It can also be observed that both FAOA and ReSC fail to notice the critical word with sentence level embedding, but our proposed method correctly localizes the referent.

We further visualize the cross-modal attention weights between individual word tokens and the pixel-level visual feature in Fig. \ref{fig:amap}. It can be observed that the same word token from two sentences attends to pixels in similar locations (1,2,3,5,6th columns of the first sub-row), but the changed keywords from two sentences (4th column of the first sub-row) have different focuses in attention maps in the first decoder layer (first sub-row). As the inference goes deeper via decoder layers (2nd and 3rd sub-rows), this different focus will guide other word tokens to focus on the correct referent (\textit{cat under chair} and \textit{cat on chair}) via language-vision feature fusion. 

\begin{figure*}[!htb]
\centering
\resizebox{\textwidth}{!}{
\begin{tabulary}{\linewidth}{c c c c c c c}
\HUGE \textbf{(a)} &
\raisebox{-.5\height}{\includegraphics{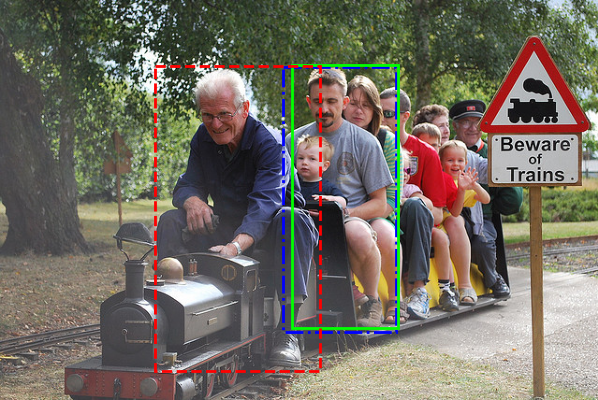}} & \raisebox{-.5\height}{\includegraphics{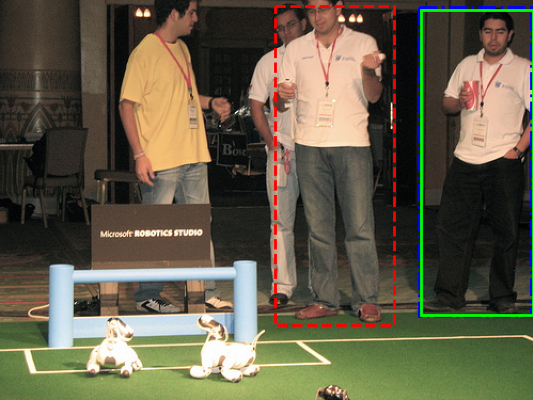}} &
\raisebox{-.5\height}{\includegraphics{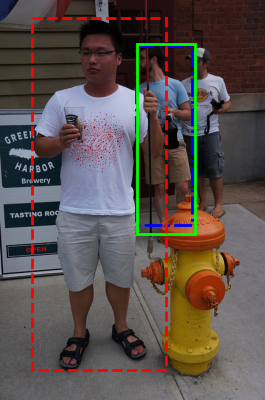}} & 
\raisebox{-.5\height}{\includegraphics{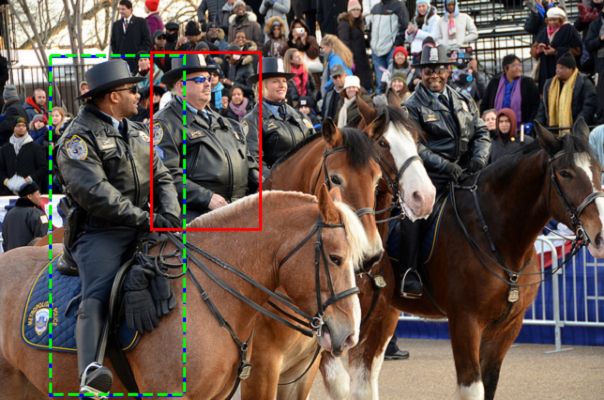}} & \raisebox{-.5\height}{\includegraphics{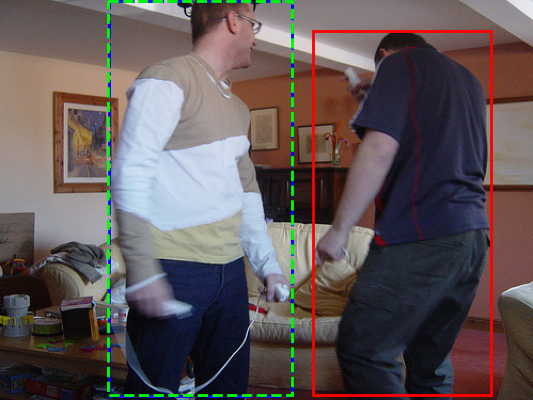}} \\\\
 & 
\textit{\thead{Q:dude in gray t shirt}}& 
\textit{\thead{Q:man holding drink black pants}} & 
\textit{\thead{Q:blue shirt\\ tan shorts}}&
\textit{\thead{Q:police on horse with turned head}}&
\textit{\thead{Q:weird shirt}}\\\\

\HUGE \textbf{(b)} & 
\raisebox{-.5\height}{\includegraphics{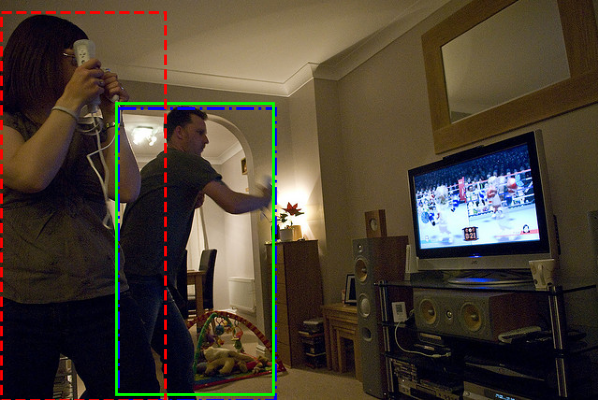}} & \raisebox{-.5\height}{\includegraphics{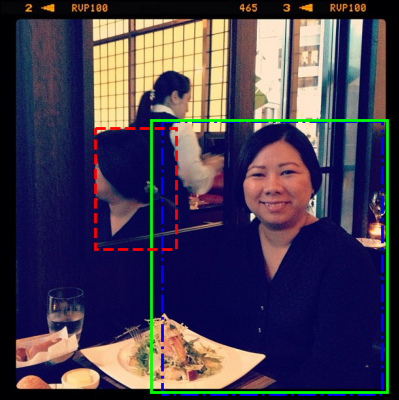}} &
\raisebox{-.5\height}{\includegraphics{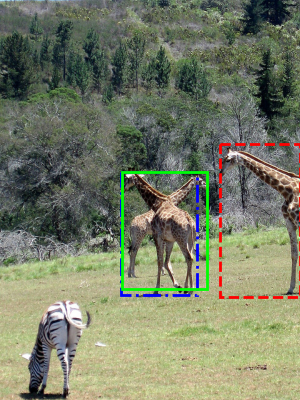}} & 
\raisebox{-.5\height}{\includegraphics{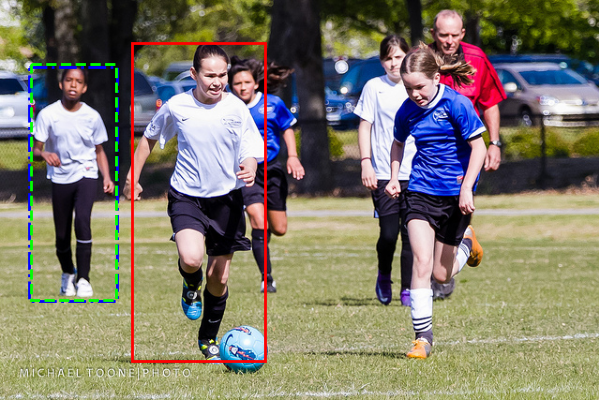}} & \raisebox{-.5\height}{\includegraphics{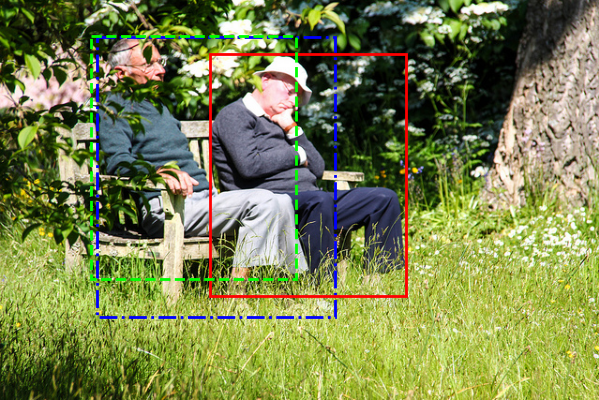}} \\\\
 & 
\textit{\thead{Q:man}}& 
\textit{\thead{Q:seated woman}} & 
\textit{\thead{Q:2 giraffes standing\\ near each other, \\looking in\\opposite directions}}&
\textit{\thead{Q:child wearing black pants}}&
\textit{\thead{Q:a man with a green sweater\\and gray pants with his hand\\resting on the bench's armrest}}\\\\
\end{tabulary}
}
\vspace*{-3mm}
\caption{Qualitative example comparisons with main stream two-stage methods. The ground truth boxes are denoted in dotted dashed line. Predictions from our method are with solid line and compared methods are denoted with dashed line. The color code for all methods: green for correct prediction ($IoU>0.5$), red for incorrect and blue for ground truth boxes. Comparison results vs MAttNet\cite{MATTNET} on REFCOCO+ are shown in first row. Second row shows comparison with Ref-NMS\cite{refnms}. For both comparisons first three images show our successful cases over other methods while last two show our failure cases.}
\label{fig:twos_comp}
\end{figure*}

Qualitative comparison examples between our method and state-of-the-art two-stage methods are shown in Fig. \ref{fig:twos_comp}. It can be seen that some of the incorrect predictions by two-stage methods for referents with color or action attributes can be corrected by our method. Nonetheless, long query sentence with complex structure is still a challenging case for our model.

\section{Conclusion and future work.}
In this paper we propose Word2Pix: a one-stage visual grounding network that enables word to pixel attention learning based on encoder-decoder transformer architecture. The proposed method outperforms state-of-the-art one-stage methods with a large margin and also surpasses existing best two-stage models while at the same time maintains the benefit of real-time inference speed via end-to-end training. Ablation experiments and qualitative examples illustrate the importance and effectiveness of word to pixel attention over holistic sentence embedding for one-stage methods in visual grounding tasks. However, slow convergence in training and how to further explore the structure of the query sentence remain as open challenges which will also be the possible direction for future work.

\ifCLASSOPTIONcaptionsoff
  \newpage
\fi

\bibliographystyle{IEEEtran}
\bibliography{my_vg}

\end{document}